\documentclass[12pt,a4paper]{elsarticle}
\usepackage{float}
\usepackage{amsmath}\allowdisplaybreaks
\usepackage{amssymb}
\usepackage{graphicx}
\usepackage[hidelinks]{hyperref}
\usepackage{breakurl}
\usepackage[T1]{fontenc}
\usepackage[english]{babel}
\usepackage[utf8]{inputenc}
\usepackage{cmap}
\usepackage{mathrsfs}
\usepackage{color}
\usepackage{multirow}
\usepackage{listings}
\definecolor{dkgreen}{rgb}{0,0.6,0}
\definecolor{gray}{rgb}{0.5,0.5,0.5}
\definecolor{mauve}{rgb}{0.58,0,0.82}
\let\origthelstnumber\thelstnumber%
\makeatletter%
\newcommand*\Suppressnumber{%
  \lst@AddToHook{OnNewLine}{%
    \let\thelstnumber\relax%
     \advance\c@lstnumber-\@ne\relax%
    }%
}%
\newcommand*\Reactivatenumber{%
  \lst@AddToHook{OnNewLine}{%
   \let\thelstnumber\origthelstnumber%
   \advance\c@lstnumber\@ne\relax}%
}%
\newfloat{lstfloat}{htbp}{lop}%
\floatname{lstfloat}{Listing}%
\newcommand*\patchenvironment[1]{%
\expandafter\let\csname old#1\expandafter\endcsname\csname #1\endcsname
\expandafter\let\csname oldend#1\expandafter\endcsname\csname end#1\endcsname
\renewenvironment{#1}%
{\linenomath\csname old#1\endcsname}%
{\csname oldend#1\endcsname\endlinenomath}%
}%
\usepackage{tikz}
\usepackage{scalerel}
\usetikzlibrary{svg.path}
\definecolor{orcidlogocol}{HTML}{A6CE39}
\tikzset{
    orcidlogo/.pic={
        \fill[orcidlogocol] svg{M256,128c0,70.7-57.3,128-128,128C57.3,256,0,198.7,0,128C0,57.3,57.3,0,128,0C198.7,0,256,57.3,256,128z};
        \fill[white] svg{M86.3,186.2H70.9V79.1h15.4v48.4V186.2z}
        svg{M108.9,79.1h41.6c39.6,0,57,28.3,57,53.6c0,27.5-21.5,53.6-56.8,53.6h-41.8V79.1z M124.3,172.4h24.5c34.9,0,42.9-26.5,42.9-39.7c0-21.5-13.7-39.7-43.7-39.7h-23.7V172.4z}
        svg{M88.7,56.8c0,5.5-4.5,10.1-10.1,10.1c-5.6,0-10.1-4.6-10.1-10.1c0-5.6,4.5-10.1,10.1-10.1C84.2,46.7,88.7,51.3,88.7,56.8z};
    }
}
\newcommand\orcidicon[1]{\href{https://orcid.org/#1}{\mbox{\scalerel*{
                \begin{tikzpicture}[yscale=-1,transform shape]
                \pic{orcidlogo};
                \end{tikzpicture}
            }{|}}}}
\newcommand{\eg}{\emph{e}.\emph{g}.,~} %
\newcommand{\ie}{\emph{i}.\emph{e}.,~}%
\newcommand{\onto}[1]{\text{{\fontfamily{lmss}\selectfont {#1}}}}%
\newcommand{\NL}[1]{\begin{tabular}[c]{@{}c@{}}#1\end{tabular}}%
\newcommand{\NLt}[1]{\begin{tabular}[t]{@{}l@{}}#1\end{tabular}}%
\newcommand{\DESCR}[1]{\color{dkgreen}$\mathcal{E}{:}\,\{\onto{#1}\}$}%
\newcommand{\LCOM}[1]{\color{dkgreen}#1}%
\newcommand{\FUNC}[1]{\texttt{#1}:~}%
\newcommand{\FullClassDescriptor}{\href{https://github.com/TheEngineRoom-UniGe/OWLOOP/blob/master/src/main/java/it/emarolab/owloop/descriptor/utility/classDescriptor/FullClassDesc.java}{\texttt{FullClassDescriptor}}}
\newcommand{\FullIndividualDescriptor}{\href{https://github.com/TheEngineRoom-UniGe/OWLOOP/blob/master/src/main/java/it/emarolab/owloop/descriptor/utility/individualDescriptor/FullIndividualDesc.java}{\texttt{FullIndividualDescriptor}}}
\newcommand{\FullObjectPropertyDescriptor}{\href{https://github.com/TheEngineRoom-UniGe/OWLOOP/blob/master/src/main/java/it/emarolab/owloop/descriptor/utility/objectPropertyDescriptor/FullObjectPropertyDesc.java}{\texttt{FullObjectPropertyDescriptor}}}
\newcommand{\FullDataPropertyDescriptor}{\href{https://github.com/TheEngineRoom-UniGe/OWLOOP/blob/master/src/main/java/it/emarolab/owloop/descriptor/utility/dataPropertyDescriptor/FullDataPropertyDesc.java}{\texttt{FullDataPropertyDescriptor}}}
\restylefloat{table}
\restylefloat{figure}
\makeatletter
\def\ps@pprintTitle{%
 \let\@oddhead\@empty
 \let\@evenhead\@empty
 \def\@oddfoot{}%
 \let\@evenfoot\@oddfoot}
\makeatother
\journal{}
\hypersetup
{
    pdfauthor={Luca Buoncompagni, Syed Yusha Kareem, and Fulvio Mastrogiovanni},
    pdfsubject={Elsevier SoftwareX publications},
    pdftitle={OWLOOP: A Modular API to Describe OWL Axioms in OOP Objects Hierarchies},
    pdfkeywords={Object-Oriented Programming (OOP), Ontology Web Language (OWL), Application Programming Interface (API), Software Architecture}
}
\begin{document}
\begin{frontmatter}
\title{OWLOOP: A Modular API to Describe\\OWL Axioms in OOP Objects Hierarchies$^\star$}
\author[1,2]{Luca~Buoncompagni\corref{equal}\textsuperscript{\orcidicon{0000-0001-8121-1616},}}%
\author[1]{Syed~Yusha~Kareem\corref{equal}\textsuperscript{\orcidicon{0000-0002-2360-1680},}}%
\author[1]{and~Fulvio~Mastrogiovanni\textsuperscript{\orcidicon{0000-0001-5913-1898},}}%
\cortext[equal]{These authors contributed equally to this work.
\\\indent~~%
\emph{Email addresses}: 
\href{mailto:luca.buoncompagni@edu.unige.it}{luca.buoncompagni@edu.unige.it} (Luca Buoncompagni),\\
\href{mailto:kareem.syed.yusha@dibris.unige.it}{kareem.syed.yusha@dibris.unige.it} (Syed Yusha Kareem), and\\ 
\href{mailto:fulvio.mastrogiovanni@unige.it}{fulvio.mastrogiovanni@unige.it} (Fulvio Mastrogiovanni).}
\address[1]{Department of Informatics, Bioengineering, Robotics and Systems Engineering, University of Genoa, Via Opera Pia 13, 16145, Genoa, Italy.}
\address[2]{Teseo s.r.l., Piazza Montano 2a, 16151, Genoa, Italy.}
\begin{abstract}%
OWLOOP is an Application Programming Interface (API) for using the Ontology Web Language (OWL) by the means of Object-Oriented Programming (OOP).
It is common to design software architectures using the OOP paradigm for increasing their modularity.
If the components of an architecture also exploit OWL ontologies for knowledge representation and reasoning, they would require to be interfaced with OWL axioms. 
Since OWL does not adhere to the OOP paradigm, such an interface often leads to boilerplate code affecting modularity, and OWLOOP is designed to address this issue as well as the associated computational aspects.
We present an extension of the OWL-API to provide a general-purpose interface between OWL axioms subject to reasoning and modular OOP objects hierarchies.
\\[.7em]
\textbf{Keywords}\\[.2em]
Object-Oriented Programming (OOP) \sep 
Ontology Web Language (OWL) \sep 
Application Programming Interface (API) \sep 
Software Architecture.
\\[1em]
{
\indent
$^\star$ 
\small
This manuscript has been submitted to the \href{https://www.journals.elsevier.com/softwarex?producttype=journals}{SoftwareX Elsevier journal} on the 12th of January 2021, revised on the 18th of November 2021, accepted on the 14th of December 2021, and published on the 30th of December 2021.
This document contains the published version of the manuscript. Please, cite this work as:
\begin{center}
\vspace{-.1em}
\noindent
\begin{minipage}{.8\textwidth}
\small
Luca Buoncompagni, Syed Yusha Kareem, and Fulvio Mastrogiovanni, 
``OWLOOP: A Modular API to Describe OWL Axioms in OOP Objects Hierarchies'', 
SoftwareX, January 2022, 100952, Vol. 17, Elsevier, 
\url{https://doi.org/10.1016/j.softx.2021.100952}.    
\end{minipage}    
\end{center}
}
\vspace{-1.2em}
\end{abstract}

\end{frontmatter}

\vfill~%

\section*{Code Metadata}
\label{}
\begin{table}[H]
    \centering
	\small
	\begin{tabular}{|l|p{.4\textwidth}|p{.47\textwidth}|}
		\hline
		C1 & Current code version & v2.1\\
		\hline
		C2 & Permanent link to code/repository used for this code version & \url{https://github.com/TheEngineRoom-UniGe/OWLOOP}\\
		\hline
		C4 & Legal Code License & GNU General Public License v3.0\\
		\hline
		C5 & Code versioning system used & git\\
		\hline
		C6 & Software code languages, tools, and services used & code language: Java\\
		\hline
		C7 & Compilation requirements, operating environments \& dependencies & \NLt{Java~v1.8.0, Gradle~v5.2.1,\\Junit~v4.12, aMOR~v2.2,\\OWL-API~v5.0.5, openllet~v2.5.1}\\
		\hline
		C8 & Link to developer documentation/manual & \url{github.com/TheEngineRoom-UniGe/OWLOOP/wiki}\\
		\hline
		C9 & Support email for questions & \NLt{\href{mailto:luca.buoncompagni.unige@gmail.com}{luca.buoncompagni.unige@gmail.com},\\\href{mailto:kareem.syed.yusha@dibris.unige.it}{kareem.syed.yusha@dibris.unige.it}}\\
		\hline
	\end{tabular}
	\caption{Code metadata}
	\label{tab:metadata}
\end{table}

\section{Motivation and Significance}
\label{sec:Motiv and Sig}
\noindent
The Web Ontology Language (OWL) is a semantic language standardised by the World Wide Web Consortium (W3C)~\cite{d2004OWL}. 
It is used to formalise implicit and explicit knowledge as logic-based \emph{axioms} among symbols to represent \emph{things}, groups of things, and relations between things in a domain of interest. 
Ontologies provide the ability to reason on a semantic knowledge representation, which is particularly important for applications involving autonomous systems and human-machine interaction.

From a software architecture perspective, an ontology is meant to be used in synergy with other software components~\cite{wache2001ontology}, which are usually implemented with the Object-Oriented Programming (OOP) paradigm.
However, issues occur since OWL axioms are not structured in an ontology following the OOP formalism.
Although there are similarities between the knowledge in an ontology and OOP-based data structures (\eg \emph{classes}, \emph{objects}, or \emph{properties}) there are also non-trivial differences, an exhaustive list of which can be found in~\cite{h2006Semantic}. 
This is one of the reasons why software developers are typically reluctant to use ontologies in their architectures~\cite{s2018Objectoriented}.

As surveyed in~\cite{s2018Objectoriented}, for accessing and integrating ontologies in a software architecture, the active and passive \emph{OWL to OOP mapping} can be used.
The \emph{active} mapping transforms an ontology from its syntactic form, \eg based on the eXtensible Markup Language (XML), to code statements in a target programming language. 
With an active mapping, there is the possibility of reasoning over the executable ontology at runtime~\cite{stevenson2011sapphire,baset2016ontojit,j2017Owlready}, but this process is limited to the amount of fitness between the OWL language and its OOP counterpart.
Instead, with the \emph{passive} OWL to OOP mapping~\cite{s2018Objectoriented}, an ontology is integrated with the OOP language through an external inference engine, \ie a reasoner like Hermit~\cite{glimm2014hermit} or Pellet~\cite{sirin2007pellet}, among others. 
The reasoner exploits the \emph{factory pattern} to create immutable OOP objects containing snapshots of the knowledge in the ontology, which is an OWL-based data structure loaded in memory.

The passive mapping involves an \emph{API-based} strategy, while the active mapping concerns an \emph{ontology-oriented programming} strategy~\cite{s2018Objectoriented}.
On the one hand, the passive mapping focuses on performance but it might lead to complex and voluminous source codes. 
Furthermore, with a passive mapping, OWL axioms which are related to each other in the ontology are always represented as independent OOP objects.
Therefore, passive mapping allows to use OWL axioms through OOP objects but it does not exploit the benefits of the OOP paradigm as active mapping does.
On the other hand, we are not aware of an active mapping that supports all the reasoning mechanisms implemented by the OWL reasoners used with passive mapping.

\begin{figure}
	\centering
	\footnotesize
    \def\svgwidth{.95\linewidth}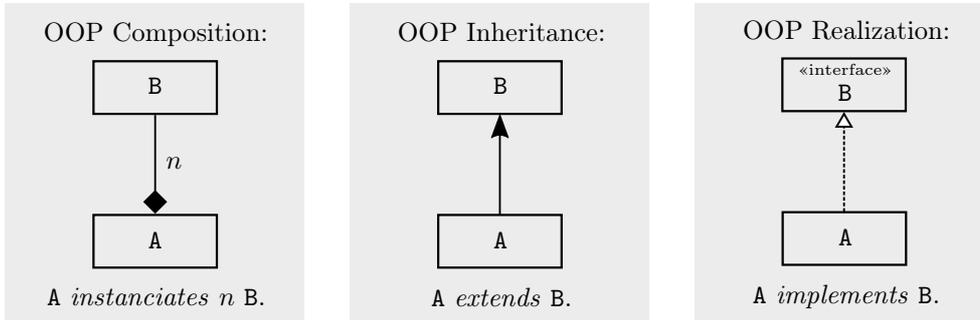
	\caption{The UML notations used in this paper.}
	\label{fig:legend}
\end{figure}

To develop a software architecture exploiting the OWL formalism, we identify two aspects concerning the design and implementation of
($i$)~the OWL-based knowledge representation from data and prior semantics, and
($ii$)~the interfaces among the components of the software architecture and OWL-based knowledge.
The former aspect can be tackled with different ontology design patterns\footnote{A collection of state of the art ontology design patterns is available at \url{http://ontologydesignpatterns.org/wiki/Ontology_Design_Patterns_._org_(ODP)}.}, which can be implemented using available tools, \eg OWL-API~\cite{m2011OWL}, Jena-API~\cite{JenaAPI} or the Prot\'eg\'e  editor~\cite{musen2015protege}.
These tools exploit string-based identifiers of OWL symbols, and they allow to define the procedures that retrieve specific knowledge through queries.
Although these tools support well the ontology design process, they might not be suitable for the latter aspect since it has different requirements, which include modularity, maintainability and synchronisation of the software architecture.

Both aspects might be tackled by developers with different expertise, \ie on OWL formalism and software architecture solutions, respectively, and they require a constant joint effort to develop an effective system.
Thus, we aim for an API which can be used to develop a modular abstraction layer interfacing ontologies and software components.
In our approach, on the one hand, experts on OWL ontologies can design semantic data representation and store it in a file as prior knowledge; also, they can develop OWLOOP-based interfaces to manipulate and query the ontology at run time.
On the other hand, the ontology and OWLOOP-based interfaces can be used by the developers focusing on architectural aspects through the OOP paradigm.

This paper presents the OWLOOP API, which implements a passive OWL to OOP mapping for integrating ontologies and software architectures.
We aim to entirely support OWL reasoners and exploit the benefits of the OOP paradigm by avoiding the factory design pattern.
The OWLOOP API interfaces OWL axioms with \emph{descriptors}, which are OOP classes representing a fragment of the ontology, but they are neither immutable nor independent.
In other words, descriptors can be designed within an OOP-based hierarchy of classes.
OWLOOP supports the implementation of general-purpose and modular OWL to OOP mapping, which can improve flexibility, reusability and maintainability of software architectures exploiting ontologies.

\begin{figure}
	\centering
    \footnotesize
    \def\svgwidth{.95\linewidth}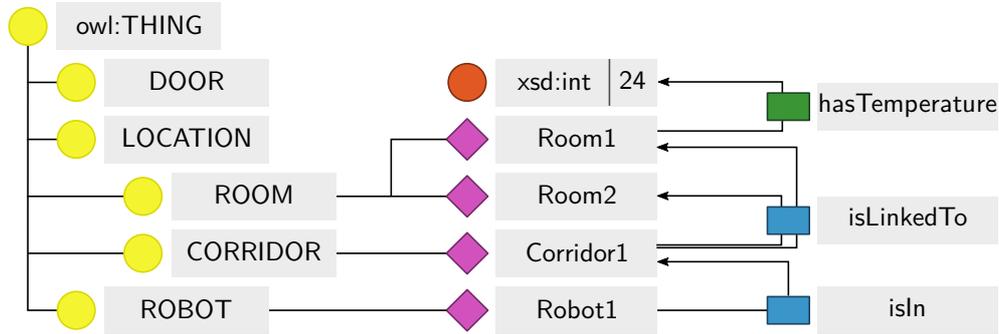
	\caption{A simple ontology used as an example throughout the paper.}
	\label{fig:owloop:example}
\end{figure}

\section{Software Description}
\label{sec:soft_desc}

\subsection{OWL Formalism Overview}
\label{sec:owlPrimer}
\noindent
Figure~\ref{fig:owloop:example} depicts an ontology used as an illustrative example throughout the paper. 
An OWL ontology is made by \emph{axioms}, \ie symbolic statements describing some \emph{entities} (listed in Table~\ref{tab:entities}) with a specific logic-based semantic expression.
Axioms are organised in three structures, \ie the Terminological and the Rule graphs (TBox and RBox), and the Assertion set (ABox).

The TBox describes concepts (\texttt{OWLClasses} or \emph{classes} in short) as a hierarchy of logic implications (\eg a \onto{ROOM} is a \emph{sub-class} of \onto{LOCATION}) with the \onto{THING} class as default root, \ie the \emph{super-class} of all possible classes, and its counterpart \onto{NOTHING}.
Instead, the ABox collects instances (\texttt{OWLNamedIndividual} or \emph{individuals}) of concepts, \eg \onto{Room1} is an \emph{instance} of the \onto{ROOM} class, which could be related among each other through roles (or \emph{properties}), \eg \onto{Room1 isLinkedTo Corridor1}.
The RBox represents properties and sub-properties similar to the TBox, and it includes specific definitions, such as \onto{transitive}, \onto{symmetric}, \emph{etc.}
A property in the Rbox can describe its \emph{inverse} property, and its \emph{range} and \emph{domain} classes, \eg \onto{hasDoor} relates instances of \onto{LOCATION} and \onto{DOOR} classes, respectively.
OWL standard distinguishes \emph{object} properties from \emph{data} properties; \texttt{OWLObjectProperty} and \texttt{OWLDataProperty}, respectively.
As introduced above, the former relates two individuals, while the latter relates an individual domain with a \emph{literal} (\texttt{OWLLiteral}) range, \eg a number.
A literal is considered as an instance of \emph{concrete} classes (\texttt{OWLDataTypes}), \eg the set of positive numbers.

OWL defines the logic \emph{disjunction} and \emph{conjunction} among classes, as well as \emph{cardinality restrictions} to represent classes through properties, \eg a \onto{CORRIDOR} is a \onto{LOCATION} that has (\ie \onto{hasDoor}) \emph{at least} 2 \onto{Doors}.
In addition, each pair of entities within the same box can be set to be \emph{equivalent} or different (\ie \emph{disjoint}), \eg \onto{Robot1} is a different individual from \onto{Room1}.
The axioms in the three boxes of the ontology can be processed by OWL-based reasoners, which check the consistency of an ontology and \emph{infer} new implicit axioms under the \emph{open word} assumption.
In addition, it can solve \emph{queries} that are given as incomplete axioms, \eg find all instances of the \onto{ROBOT} class.
To better appreciate the reasoning ability, more details of the OWL formalisms are available in \cite{motik2009owl,b2009OWL}.

To deal with such a variety of axioms semantics, the standard implementation of OWL associates each axiom $\mathcal{A}_j$ to a certain $k$-th \emph{expression} $\mathcal{E}_k$, \eg \onto{Sub}, \onto{Super}, \onto{Instance}, \onto{Equivalent}, \onto{Disjoint}, \emph{etc.}
In turn, the expression defines the types of entities that need to be specified in the axiom ${\mathcal{A}_j=\mathcal{E}_k(e_1,e_2,\ldots)}$, \eg the sub-class expression requires that $e_1$ and $e_2$ are both classes, while the instance expression requires an individual and a class.
To represent axioms suitably for OOP mapping, OWLOOP provides an $\mathcal{E}_k$ counterpart for each OWL expression, \eg an OWL \onto{ObjectPropertyAssertion} is an OWLOOP \onto{ObjectLink}, while an OWL \onto{ClassAssertion} is an OWLOOP \onto{Type}.
This paper presents OWLOOP expressions as shown in Table~\ref{tab:mapping} and introduced above, \eg some OWL axioms in our illustrative ontology are

\begin{equation}
    \begin{aligned}
        \onto{Super}                 &\big(   \onto{ROOM},           \; \onto{LOCATION}                                 \big),\\ 
        \onto{Instance}              &\big(   \onto{Room1},          \; \onto{ROOM}                                     \big),\\
        \onto{ObjectLink}            &\big(   \onto{isLinkedTo},     \; \onto{Room1}, \; \onto{Corridor1}               \big),\\
        \onto{DataLink}              &\big(   \onto{hasTemperature}, \; \onto{Room1}, \; 24                             \big),\\    
        \onto{EquivalentRestriction} &\big(   \onto{CORRIDOR},       \; \onto{min 2}, \; \onto{hasDoor}, \; \onto{DOOR} \big).
    \end{aligned}
    \label{eq:OWL}
\end{equation} 

\noindent
The OWLOOP API enables interaction with OWL entities in an ontology by using \emph{descriptors}, \ie Java classes encapsulating reusable code to exploit OWL axioms, \eg string-based identifiers of OWL symbols and procedures based on queries.
The next Section presents the structure of OWL axioms in a descriptor, while Section~\ref{sec:soft_func} focuses on the descriptor functionalities to assert and retrieve knowledge subject to reasoning.

\subsection{Software Architecture}
\label{sec:soft_arch}
\noindent
The OWL to OOP mapping implemented by OWLOOP stores the axioms involving an $\mathcal{E}_k$ expression into a data structure
${D_k=\langle x, Y_k\rangle}$.
In it, $x$ is an OWL entity called \emph{ground}, and 
${Y_k=k{:}[\{y_{1}\},\{y_{2}\},\ldots]}$ is an \emph{entity set} such that the OWL axiom
${\mathcal{E}_k(x,y_{i})}$ 
is derived for each $y_i\in Y_k$. 
For example, we encode the OWL axioms in \eqref{eq:OWL} as
\begin{equation}
    \begin{aligned}
        D_1 &= \big\langle \onto{ROOM},\;\;   \onto{Super}{:}                 \big[ \{\onto{LOCATION}\},\;\;\{\onto{THING}\}         \big]\big\rangle,\\
        D_2 &= \big\langle \onto{ROOM},\;\;   \onto{Instance}{:}              \big[ \{\onto{Room1}\},\;\{\onto{Room2}\}              \big]\big\rangle,\\
        D_3 &= \big\langle \onto{Room1},\;\;  \onto{ObjectLink}{:}            \big[ \{\onto{isLinkedTo},\;\onto{Corridor1}\}         \big]\big\rangle,\\
        D_4 &= \big\langle \onto{Room1},\;\;  \onto{DataLink}{:}              \big[ \{\onto{hasTemperature},\;\onto{24}\}            \big]\big\rangle,\\
        D_5 &= \big\langle \onto{CORRIDOR},\; \onto{EquivalentRestriction}{:} \big[ \{\onto{min}\,2,\;\onto{hasDoor},\;\onto{DOOR}\} \big]\big\rangle.
    \end{aligned}
    \label{eq:OWLOOP}
\end{equation}
Each $i$-th element of the entity set $Y_k$ is an OWLOOP entity, which can either be an OWL entity (as in $D_1$ and $D_2$) or a structure of OWL entities, \ie as specified in Table~\ref{tab:entities}, an \texttt{OWLOOPObject} (in $D_3$), an \texttt{OWLOOPData} (in $D_4$) or an \texttt{OWLOOPRestriction} (in $D_5$).

\begin{table}
    \renewcommand*{\arraystretch}{1.05}
    \footnotesize
    \centering
    \begin{minipage}{.25\textwidth}
        \begin{tabular}{@{\,}c@{\,}}                \hline
            \texttt{OWLEntity}           \\      \hline
            \texttt{OWLClass},           \\[.2em]
            \texttt{OWLDataType},        \\[.2em]
            \texttt{OWLNamedIndividual}, \\[.2em]
            \texttt{OWLLiteral},         \\[.2em]
            \texttt{OWLObjectProperty},  \\[.2em]
            \texttt{OWLDataProperty}.    \\      \hline                      
        \end{tabular} 
    \end{minipage}
    \hfill%
    \begin{minipage}{.7\textwidth}
        \begin{tabular}{@{\,}r@{~}l@{\,}}\hline
            \multicolumn{2}{c}{\texttt{OWLOOPEntity} \emph{(extends} \texttt{OWLEntity}\emph{)}}\\\hline
            \texttt{OWLOOPObject}:  & $\big\langle\texttt{OWLObjectProperty}, \texttt{OWLNamedIndividual}\big\rangle$,       \\[.5em]
            \texttt{OWLOOPData}:    & $\big\langle\texttt{OWLDataProperty}, \texttt{OWLLiteral}\big\rangle$,                 \\[.5em]
            \multirow{4}{*}{\begin{tabular}[c]{@{}c@{}}\texttt{OWLOOP}\\\texttt{Restriction}:\end{tabular}} 
                                    & $\big\langle\texttt{OWLClass}\big\rangle$, or                                          \\
                                    & $\big\langle\mathcal{C}, \texttt{OWLClass}\big\rangle$, or                             \\
                                    & $\big\langle\mathcal{C}, \texttt{OWLObjectProperty}, \texttt{OWLClass}\big\rangle$, or \\
                                    & $\big\langle\mathcal{C}, \texttt{OWLDataProperty}, \texttt{OWLDataType}\big\rangle$.   \\\hline
        \end{tabular}
    \end{minipage}
    \vspace{.5em}
    \caption{%
        On the left--hand side, the possible types of entities in an OWL ontology. 
        OWLOOP entities can either be OWL entities or structures of OWL entities, which are shown on the right.
        The cardinality $\mathcal{C}$ spans in an OWL axiom among $\{\onto{some},\;\onto{only},\;\onto{min}\,n,\;\onto{max}\,n,\;\onto{exact}\,n\}$ where ${n\in\mathbb{N}^+}$.
    }%
    \label{tab:entities}%
\end{table}

\begin{table}
    \renewcommand*{\arraystretch}{1.05}
    \footnotesize
    \centering
    \begin{tabular}{c|c|c}\hline
    \multicolumn{1}{c|}{\textit{Ground ($x$)}} & \multicolumn{1}{c|}{\textit{Expression ($\mathcal{E}_k$)}}  & \multicolumn{1}{c}{\textit{Entity set element ($y_i$)}} \\ \hline
        \NL{\texttt{OWLClass}\\[.5em](\texttt{Class}\\\texttt{Descriptor})}   
            & \NL{\onto{Equivalent}\\\onto{Disjoint}\\\onto{Super}\\\onto{Sub}\\\onto{Instance}\\\onto{EquivalentRestriction}}
            & \NL{\texttt{OWLClass}\\\texttt{OWLClass}\\\texttt{OWLClass}\\\texttt{OWLClass}\\\texttt{OWLNamedIndividual}\\\texttt{OWLOOPRestriction}}
        \\\hline
        \NL{\texttt{OWLNamedIndividual}\\[.5em](\texttt{Individual}\\\texttt{~Descriptor})}                  
            & \NL{\onto{Type}\\\onto{Equivalent}\\\onto{Disjoint}\\\onto{ObjectLink}\\\onto{DataLink}}
            & \NL{\texttt{OWLClass}\\\texttt{OWLNamedIndividual}\\\texttt{OWLNamedIndividual}\\\texttt{OWLOOPObject}\\\texttt{OWLOOPData}}
        \\\hline
        \NL{\texttt{OWLObjectProprty}\\[.5em](\texttt{ObjectProperty}\\\texttt{~Descriptor})}                     
            & \NL{\onto{Equivalent}\\\onto{Disjoint}\\\onto{Sub}\\\onto{Super}\\\onto{Inverse}\\\onto{Domain}\\\onto{Range}}
            & \NL{\texttt{OWLObjectProperty}\\\texttt{OWLObjectProperty}\\\texttt{OWLObjectProperty}\\\texttt{OWLObjectProperty}\\\texttt{OWLObjectProperty}\\\texttt{OWLOOPRestriction}\\\texttt{OWLOOPRestriction}}
        \\\hline
        \NL{\texttt{OWLDataProperty}\\[.5em](\texttt{DataProperty}\\\texttt{~Descriptor})}
            & \NL{\onto{Equivalent}\\\onto{Disjoint}\\\onto{Sub}\\\onto{Super}\\\onto{Domain}\\\onto{Range}}
            & \NL{\texttt{OWLDataProperty}\\\texttt{OWLDataProperty}\\\texttt{OWLDataProperty}\\\texttt{OWLDataProperty}\\\texttt{OWLOOPRestriction}\\\texttt{OWLOOPRestriction}}\\\hline
    \end{tabular}
    \vspace{.5em}
    \caption{%
        The expressions that OWLOOP can currently map into axioms (in the second column), organised by the four types of descriptors, \ie class, individual, object and data property descriptors.
        The type of each descriptor is given by the type of its ground entity (in the first column), and it can describe the shown expressions, each with a specific entity set (third column).
    }%
    \label{tab:mapping}
\end{table}

\begin{figure}
	\scriptsize
	\centering
	\def\svgwidth{.95\linewidth}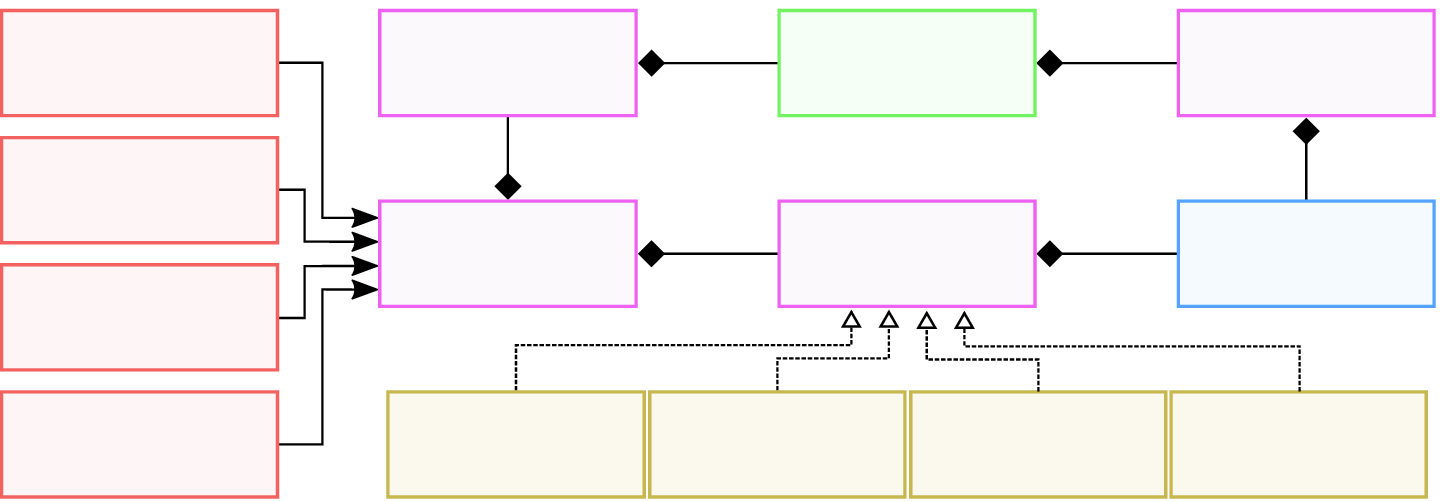
	\\[.5em]
	\def\svgwidth{.95\linewidth}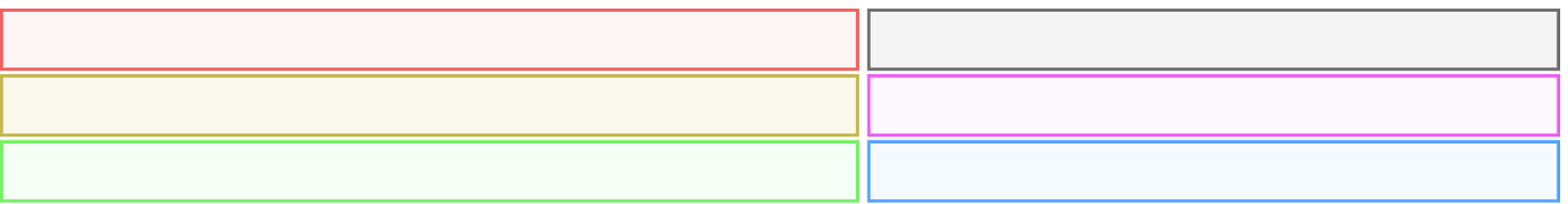
	\caption{The general-purpose definition of OWLOOP \texttt{Descriptor}. Colours identify OOP packages, which are also consistently used in Figures~\ref{fig:classUML}--\ref{fig:objectpropertyUML}.}
	\label{fig:coreUML}
\end{figure}

OWLOOP represents the data structure $D$ in the OOP interface named \texttt{Descriptor}, which is implemented as shown in Figure~\ref{fig:coreUML} through the Unified Modelling Language (UML), whose notation is summarised in Figure~\ref{fig:legend}.
From~\eqref{eq:OWLOOP} it is possible to deduce that descriptors with the same type of ground might be merged to encode different axioms, \eg $D_1$ and $D_2$ can be represented in a \emph{compound} descriptor ${D_c=\langle x, Y_1, Y_2\rangle}$,
where $Y_1$ and $Y_2$ concern the \onto{Super} and \onto{Instance} expressions.
Hence, OWLOOP provides four different descriptors based on their ground, \ie \texttt{ClassDescriptor}, \texttt{IndividualDescriptor}, \texttt{ObjectPropertyDescriptor}, and \texttt{DataPropertyDescriptor}.

In order to address an application, developers should design suitable compound descriptors with sets of axioms based on Table~\ref{tab:mapping}.
Four steps should be followed to design a compound descriptor based on the functionalities that an OOP class can inherit from OWLOOP interfaces, \eg as shown in Listing~\ref{lst:descrImpl}:
($i$)~assign a ground entity by inheriting from one of the interfaces provided by the 
$^{\star}\!$\href{https://github.com/TheEngineRoom-UniGe/OWLOOP/tree/master/src/main/java/it/emarolab/owloop/descriptor/construction/descriptorGround}{\texttt{.descriptorGround}} 
package;
($ii$)~inherit from extensions of the \texttt{Descriptor} interfaces (available in the package 
$^{\star}\!$\href{https://github.com/TheEngineRoom-UniGe/OWLOOP/tree/master/src/main/java/it/emarolab/owloop/descriptor/construction/descriptorExpression}{\texttt{.descriptorExpression}}) 
the representation of some expressions $\mathcal{E}_k$;
($iii$)~consistently for each $\mathcal{E}_k$ expression, instantiate an empty entity set based on the 
$^{\star}\!$\href{https://github.com/TheEngineRoom-UniGe/OWLOOP/tree/master/src/main/java/it/emarolab/owloop/descriptor/construction/descriptorEntitySet}{\texttt{.descriptorEntitySet}} package;
($iv$)~for each $\mathcal{E}_k$, specify a descriptor to \emph{build} (addressed in the next Section).

\begin{figure}
	\scriptsize
	\centering
	\def\svgwidth{.95\linewidth}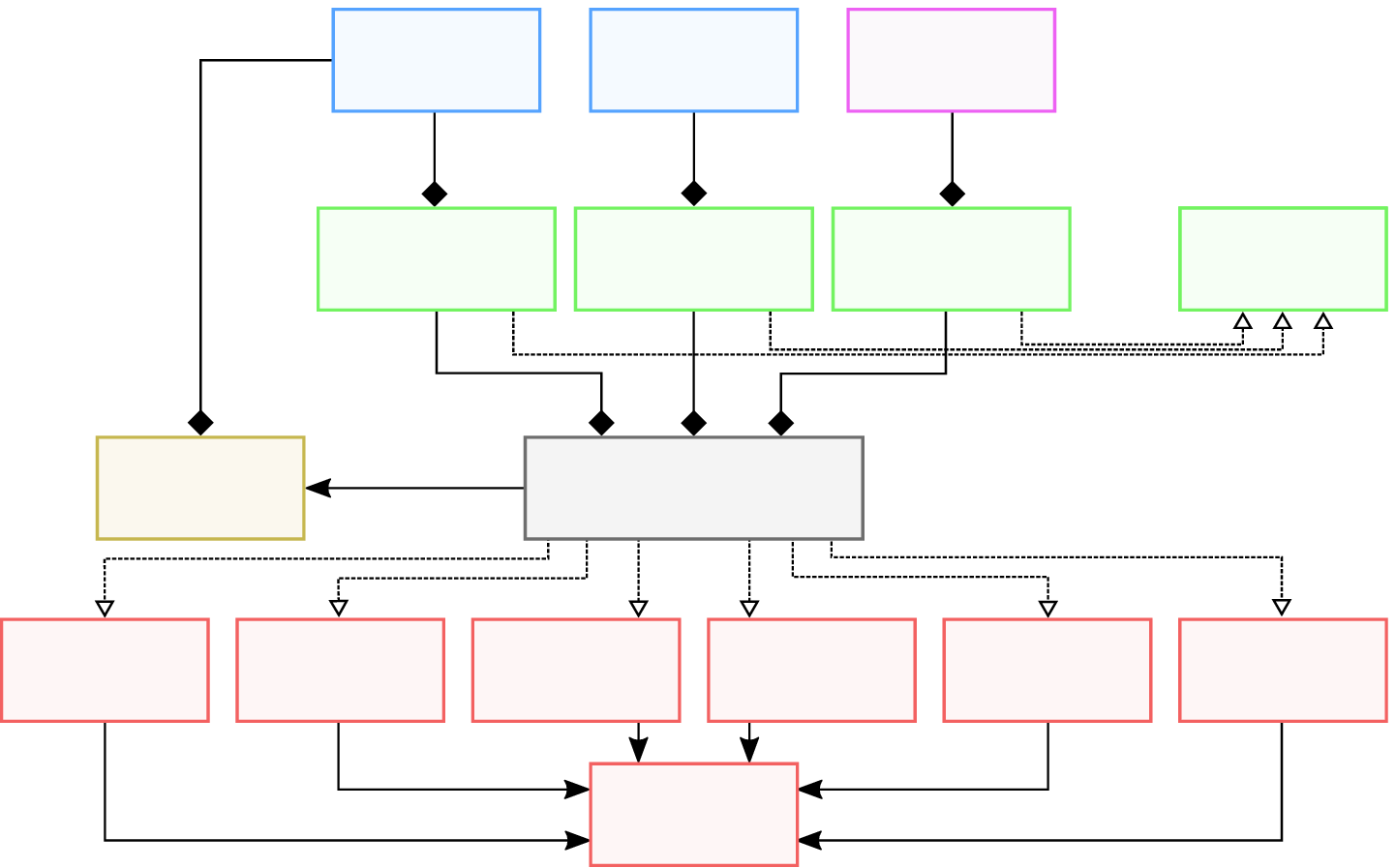
	\caption{The implementation of the \FullClassDescriptor.}
	\label{fig:classUML}
\end{figure}

\begin{figure}
	\centering
	\scriptsize
	\def\svgwidth{.95\linewidth}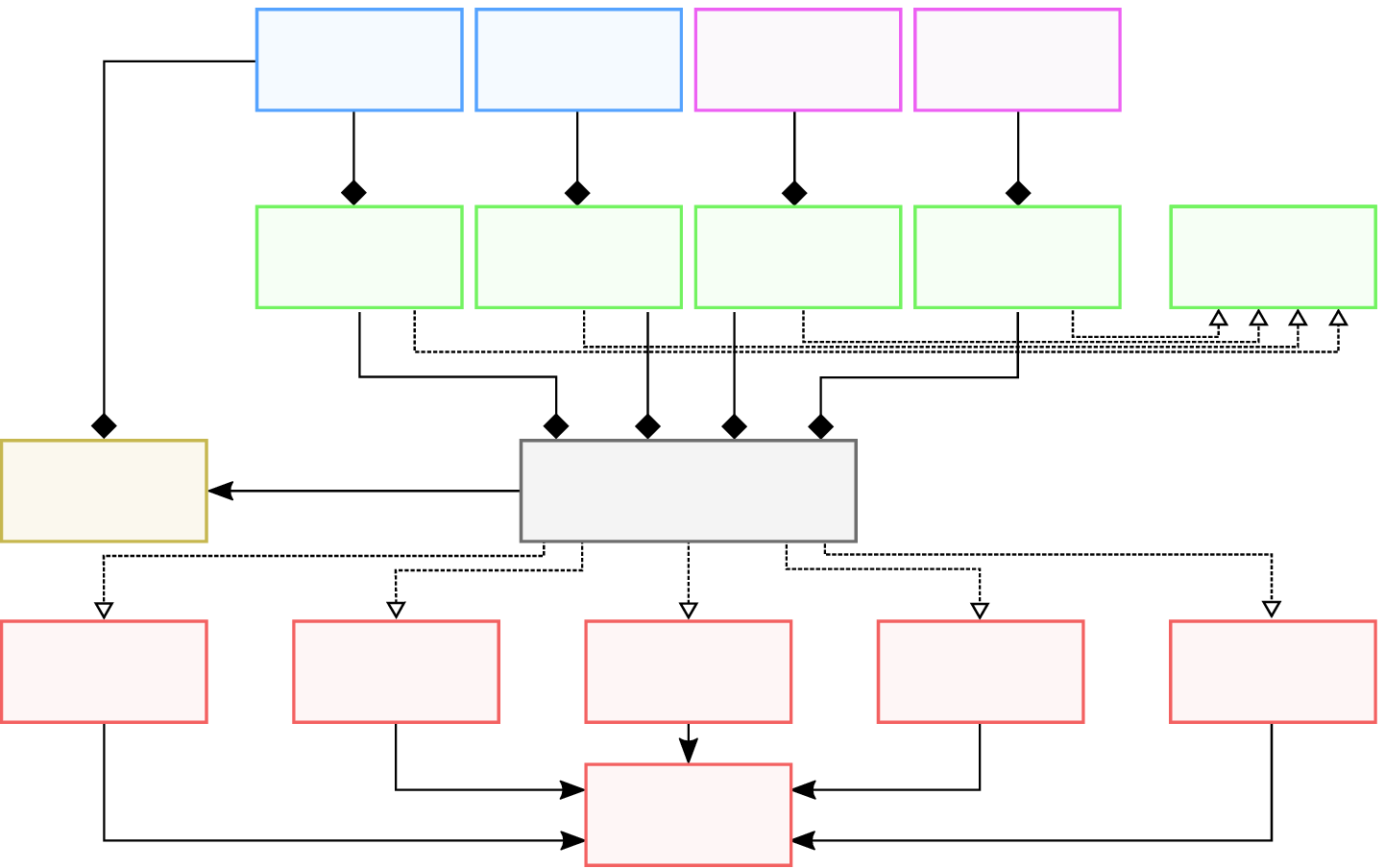
    \caption{The implementation of the \FullIndividualDescriptor.}
	\label{fig:individualUML}
\end{figure}

\begin{figure}
	\centering
	\scriptsize
	\def\svgwidth{.95\linewidth}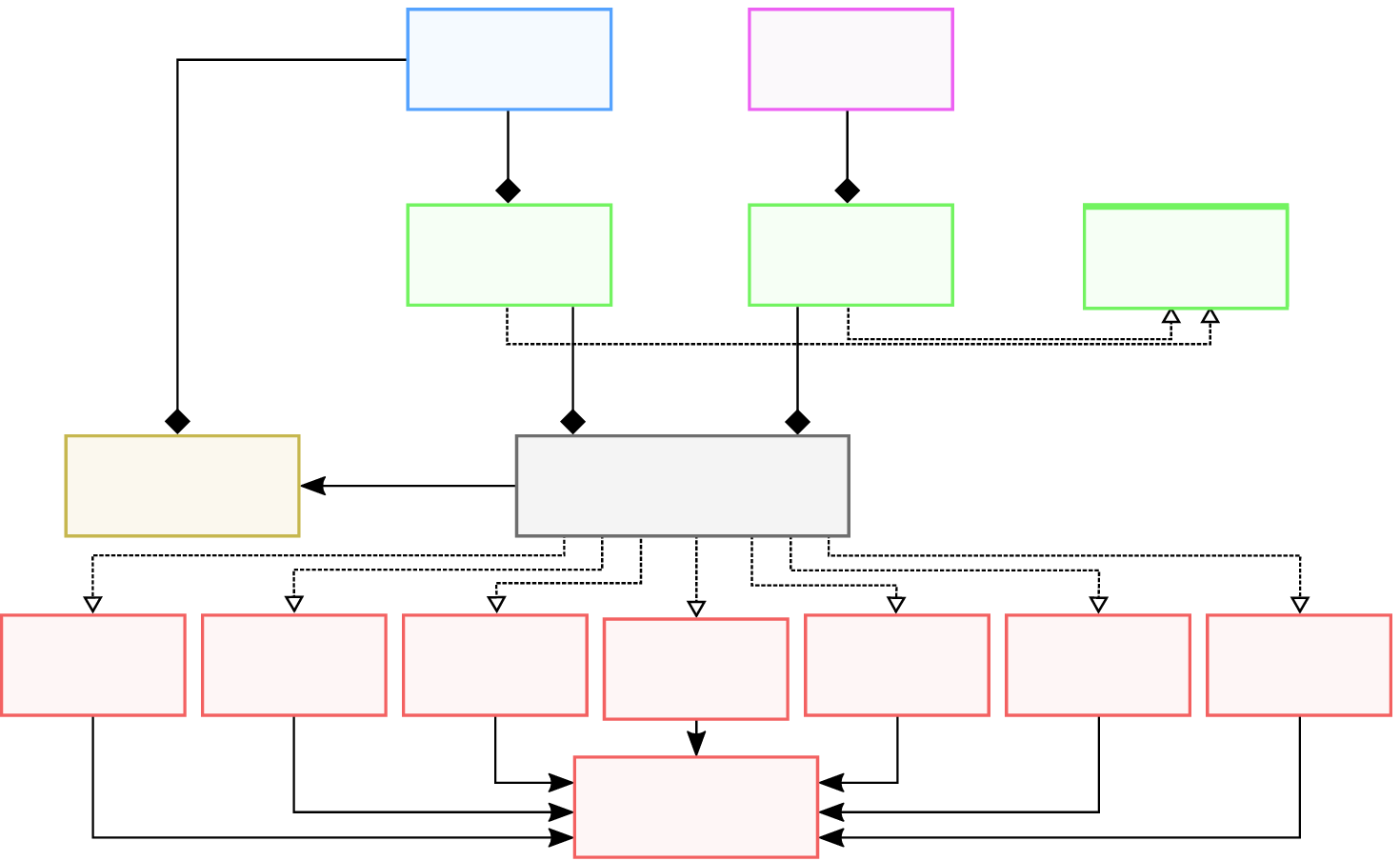
    \caption{The implementation of the \FullObjectPropertyDescriptor.}
	\label{fig:objectpropertyUML}
\end{figure}

Figure~\ref{fig:classUML} shows the implementation of the \FullClassDescriptor, which is a compound descriptor involving all the class expressions shown in the first row of Table~\ref{tab:mapping}.
Therefore it requires as \texttt{EntitySet} four sets of \texttt{Classes}, a set of \texttt{Individuals} and a set of \texttt{Restrictions}.
Remarkably, any compound descriptors with a \texttt{ClassGround} that can be implemented with OWLOOP concerns a subset of expressions involved in the \FullClassDescriptor.
Figures~\ref{fig:individualUML} and \ref{fig:objectpropertyUML} respectively show the implementation of the \FullIndividualDescriptor,  and the \FullObjectPropertyDescriptor, 
from which it is possible to derive the implementation of the \FullDataPropertyDescriptor.
Indeed, since our mapping exploits the same structure $D$ for all OWL axioms, OWLOOP always relies on the same pattern to address different OWL expressions.
Thus, OWLOOP is modular, and this also facilitates the implementation of the OWL axioms not considered in the current version, as discussed in Section~\ref{sec:concl}.

\subsection{Software Functionalities}
\label{sec:soft_func}

\noindent
A descriptor derived from Table~\ref{tab:mapping} can be instantiated given 
($i$)~a ground, \ie the identifier to an OWL entity (\eg ``Corridor1''), and 
($ii$)~the reference to an ontology as required by aMOR\footnote{OWLOOP is interfaced with the ontology through the aMOR library, which wraps the OWL-API to provide helper classes. We discuss the portability of aMOR in Section~\ref{sec:impact}.}, \ie an \texttt{OntologyReference}.
For instance, Listing~\ref{lst:instanDesc} shows at Line~\ref{ln:newOnto} a function to create the ontology with an associated reasoner, which should be manually invoked.
As an example, Line~\ref{ln:instnciateDescriptor} constructs a \texttt{LinkIndividualDescr}, which is a compound descriptor grounded on \onto{Corridor1} and it is concerned with the expressions $\mathcal{E}_k$ inherited from the descriptors addressing \onto{ObjectLink} and \onto{DataLink}. 

More generally, an OOP interface \texttt{D} that extends \texttt{Descriptor} for an expression $\mathcal{E}_k$ inherits the OOP \emph{methods} addressed in the following paragraphs.
These functionalities are based on the \emph{internal state} of a descriptor (\ie the ground $x$ and the entity set $Y_k$) that can be synchronised with the ontology.
A compound descriptor is an OOP object \texttt{cd} realising some \texttt{Descriptor} interfaces, whose functionalities can be accessed with \texttt{cd.$\mathcal{E}_k$}.
For instance, the interfaces implemented by the object \texttt{cd} returned at Line~\ref{ln:instnciateDescriptor} are accessible with \texttt{cd\onto{.ObjectLink}} or \texttt{cd\onto{.DataLink}}\footnote{In this paper, we slightly simplify the syntax. See the $^{\star}\!$\href{https://github.com/TheEngineRoom-UniGe/OWLOOP/tree/master/src/test/java/it/emarolab/owloop/articleExamples}{\texttt{.articleExamples} package} for the runnable version of the illustrated examples.}.

\FUNC{D.getGround()}
This method returns the ground entity $x$. 
For a compound descriptor, \texttt{cd.$\mathcal{E}_k$.getGround()} returns the same entity for all the implemented expressions $\mathcal{E}_k$.

\FUNC{D.getEntities()}
It returns the entity set $Y_k$ associated to the $\mathcal{E}_k$ expression.
To enable the definition of compound descriptors, each descriptor \texttt{D} defines this method with a different name, \eg \texttt{cd.getObjects()} for the \onto{ObjectLink} expression, and \texttt{cd.getEquivalentIndividuals()} for \onto{Equivalent} associated to an individual ground.
These methods are used to access entities $y_i$, as well as manipulate them, \eg through \texttt{cd.removeObject()}.

\begin{lstfloat}[!t]\begin{lstlisting}
private OntologyReference createEmptyOntology() { |\label{ln:newOnto}|
    OntologyReference.activateAMORlogging(false)|\!|; // Disable logs.
    return OntologyReference.newOWLReferencesCreatedWithPellet(  |\Suppressnumber|
        "robotAtHomeOnto", // Ontology refeference name.
        "src/test/resources/robotAtHomeOntology.owl", // File path.
        "http://www.semanticweb.org/emaroLab/robotAtHomeOntology", // IRI.
        true // Synchronize the Pellet reasoner manually.
    )|\!|; |\Reactivatenumber|
}
private LinkIndividualDesc newCorridorDescriptorr() { 
    OWLReferences ontoRef = createEmptyOntology()|\!|;
    LinkIndividualDesc cd = LinkIndividualDesc("Corridor1", ontoRef)|\!|;  |\label{ln:instnciateDescriptor}|
    return cd; // Compound descriptor concerning |\DESCR{ObjectLink, DataLink}.|
}
\end{lstlisting}
\caption{Example showing the instantiation of an  OWLOOP descriptor.}
\label{lst:instanDesc}
\end{lstfloat}    

\FUNC{D.query()}
This method returns the knowledge related to the $k$-th expression that involves the ground entity in the ontology.
It returns a set structured as $Y_k$ but it does not affect the internal state of the descriptor.
It is mainly used by the \texttt{Descriptor} interface itself.

\FUNC{D.readAxioms()}
It relies on \texttt{D.getEntities()} and \texttt{D.query()} to compare the internal state of the descriptor with the state of the ontology.
It changes the entity set of the descriptor to be equal to the ontology.
It returns a list of \emph{intents} containing all the performed changes, which can be used to recover from possible inconsistencies. 
A compound descriptor provides the method \texttt{cd.readAxioms()}, which invokes \texttt{cd.$\mathcal{E}_k$.readAxioms()} for each $k$-th expression concerned by \texttt{cd}.

\FUNC{D.writeAxioms()}
It is similar to \texttt{readAxioms()} but it changes the ontology to become equal to the entity set.

\FUNC{D.build()}
It allows to use a descriptor in an OOP manner since it returns a new compound descriptor \texttt{nd} grounded in each entity $y_i$.
While designing compound descriptors, developers should specify the type of \texttt{nd} with a ground consistent with the elements in the entity set of \texttt{D}.
Then, the entity set of the new descriptors is populated through the \texttt{nd.readAxioms()} method. 
Similarly to \texttt{D.getEntities()}, each descriptor \texttt{D} defines building methods with different names to enable the definition of compound descriptors.
Remarkably, if entities $y_i$ are not OWL entities, the \texttt{build()} method would have multiple definitions, \eg for an \onto{ObjectLink} it is possible to return descriptors \texttt{nd} grounded on an object property or on an individual. 

There are also other useful functionalities implemented by aMOR, \eg to get OWL entities or save the ontology.
Moreover, the function \texttt{ontoRef.synchroniseReasoner()} performs OWL reasoning, which involves all the OWL expressions.
Remarkably, OWL querying is a time-consuming task if reasoning is required.
Hence, the reasoning process should only be used when required, \ie to affect the results of \texttt{query()} and, consequently, the outcomes of \texttt{readAxioms()} and \texttt{WriteAxioms()}.
In addition, to be compatible with other software using ontologies, aMOR provides full access to factory-based OWL to OOP mapping implemented with OWL-API.

\section{Illustrative Examples}
\label{sec:illustrative_exp}

\subsection{Asserting and Inferring OWL Axioms}
\noindent
Listing~\ref{lst:addAxioms} shows how to use compound descriptors to manipulate an ontology by adding and removing some of the axioms shown in Figure~\ref{fig:owloop:example}.
At Line~\ref{ln:getAdd} the axiom 
$\onto{ObjectLink}(\onto{isLinkedTo}, \onto{Corridor1}, \onto{Room1})$
is added to the relative entity set, and the same operation is performed at Line~\ref{ln:shortcutAdd} with a different type of input.
At Line~\ref{ln:write1}, the ontology changes such that \onto{Corridor1} is linked to \onto{Room1} and \onto{Room2}.

Lines~\ref{ln:getAdd3}--\ref{ln:justAdd3} define axioms concerning disjoint classes, and at Line~\ref{ln:write3}, the ontology is changed to contain them, \ie
$\onto{Disjoint}(\onto{ROBOT},\onto{LOCATION})$ and $\onto{Disjoint}(\onto{ROBOT},\onto{DOOR})$.
Lines~\ref{ln:getAdd2}--\ref{ln:justAdd2} add axioms to specify the domain and range of the \onto{hasDoor} property, \ie
$\onto{Domain}(\onto{hasDoor}, \onto{LOCATION})$ and $\onto{Range}(\onto{hasDoor}, \onto{DOOR})$, 
which are applied to the ontology at Line~\ref{ln:write2}. 

Line~\ref{ln:reason1} updates the reasoner to infer new knowledge in the ontology, which is then synchronised with the internal states of the descriptors at Lines~\ref{ln:read1}--\ref{ln:read3}.
By reading axioms, we can populate the entity set with implicit knowledge required for implementing procedures that are based on queries.
Finally, Line~\ref{ln:getRemouve} shows a way to remove an element from the internal state of a descriptor, and Line~\ref{ln:apply} applies those changes to the ontology.

\begin{lstfloat}[!t]\begin{lstlisting}   
OWLReferences ontoRef = this.createEmptyOntology()|\!|;
// Assert OWL Individual Expression Axioms in the ontology.
ObjectLinkIndividualDesc corrdor1 = // |\DESCR{ObjectLink, DataLink}. \Suppressnumber|
        new ObjectLinkIndividualDesc("Corridor1", ontoRef)|\!|; |\Reactivatenumber|
corrdor1.addObject("isLinkedTo", "Room1")|\!|; // |\LCOM{Add to \onto{ObjectLink} \texttt{EtitySet}.} \label{ln:getAdd}|
corrdor1.addObject(new OWLOOPObject("isLinkedTo", "Room2")|\!|)|\!|; |\label{ln:shortcutAdd}|
corrdor1.writeAxioms()|\!|; // Synchronise changes to the ontology. |\label{ln:write1}|
// Add OWL Class Expression Axioms to the ontology.
DisjointClassDesc robotClass =  // |\DESCR{Disjoint}. \Suppressnumber|
        new RestrictionClassDesc("ROBOT", ontoRef)|\!|; |\Reactivatenumber|
robotClass.addDisjointClass("LOCATION")|\!|;  // |\LCOM{Add to \onto{Disjoint} \texttt{EntitySet}.} \label{ln:getAdd3}|
robotClass.addDisjointClass(ontoRef.getOWLClass("DOOR")|\!|)|\!|; |\label{ln:justAdd3}|
corrdor1.writeAxioms()|\!|; // Synchronise changes to the ontology. |\label{ln:write3}|
// Add OWL ObjectProperty Expression Axioms to the ontology.
DomainRangeObjectPropertyDesc hasDoor = // |\DESCR{Domain, Range}. \Suppressnumber|
        new DomainRangeObjectPropertyDesc("hasDoor", ontoRef)|\!|; |\Reactivatenumber|
hasDoor.addDomainClassRestriction("LOCATION")|\!|; // |\LCOM{Add to \onto{Domain} \texttt{EntitySet}.} \label{ln:getAdd2}|
hasDoor.addRangeClassRestriction(new OWLOOPRestriction("DOOR")|\!|)|\!|; |\label{ln:justAdd2}|
corrdor1.writeAxioms()|\!|; // Synchronise changes to the ontology. |\label{ln:write2}|
// Synchronise changes to the entity sets based on the ontology.
ontoRef.synchroniseReasoner()|\!|; // Invoke OWL reasoning. |\label{ln:reason1}|
corrdor1.readAxioms()|\!|; |\label{ln:read1}|
hasDoor.readAxioms()|\!|; 
robotClass.readAxioms()|\!|; |\label{ln:read3}|
// Remove an axiom from the internal state of a descriptor.
corridor1.removeObject("isLinkedTo")|\!|)|\!|; |\label{ln:getRemouve}|
corridor1.writeAxioms()|\!|; // Synchronise changes to the ontology. |\label{ln:apply}|
\end{lstlisting}
\caption{Example showing how to manipulate axioms through descriptors.}
\label{lst:addAxioms}
\end{lstfloat}

\begin{lstfloat}[!t]\begin{lstlisting}
// |\LCOM{Load the ontology shown in Figure~\ref{fig:owloop:example} similarly to Line~\ref{ln:newOnto}.}|
OWLReferences ontoRef = this.loadOntology()|\!|; |\label{ln:loadOnto}|
// |\LCOM{Retrieve the robot location (\texttt{robotLoc}) by assuming that it is unique.}|
LinkIndividualDesc robot1 = new LinkIndividualDesc("Robot1",ontoRef)|\!|; |\label{ln:robotDs}|
robot1.addObject("isIn", true)|\!|; // Set to read only the |\LCOM{\onto{isIn}}| property once.
robot1.readAxioms()|\!|;
OWLNamedIndividual robotLoc = robot1.getIndividualFromObjectProp("isIn")|\!|; |\label{ln:robotLoc}|
// |\LCOM{Ground an individual descriptor on \texttt{robotLoc} concerning \DESCR{Type}.}|
TypeIndividualDesc locIndiv = new TypeIndividualDesc(robotLoc, ontoRef)|\!|; |\label{ln:locDescr1} \Reactivatenumber|
locIndiv.readAxioms()|\!|; // |\LCOM{read the classes of \texttt{robotLoc}.} \label{ln:locDescr2}|
// |\LCOM{Build descriptors grounded on the classes of \texttt{robotLoc}.}|
Set<SubclassDesc> locClasses = locIndiv.buildTypes()|\!|; // |\DESCR{Sub, Super}\label{ln:build}|
for(SubclassDesc locClass : locClasses) |\label{ln:for}|
    // |\LCOM{A class of \texttt{robotLoc} that only subsumes \onto{owl:NOTHING} is a leaf.}|
    if(locClass.getdsubClasses()|\!|.size() == 1) |\label{ln:ifleaf}|
        // |\LCOM{Print, \eg "\onto{Robot1} is in \onto{Corridor1}, which is a \onto{CORRIDOR}".}|
        System.out.println(robot1.getGround() + " is in " + robotLoc |\Suppressnumber\label{ln:print}| 
                             + ", which is a " +  locClass.getGround()|\!|)|\!|; |\Reactivatenumber|
\end{lstlisting}
\caption{Example showing the usage of the descriptor \texttt{build} method.}
\label{lst:buildMethod}
\end{lstfloat}

\subsection{Accessing OWL Axioms with the OOP Paradigm}
\noindent
The example in Listing~\ref{lst:buildMethod} has the objective of finding the type of room where the robot is located, \eg \onto{CORRIDOR}.
The example shows the retrieval of the individual related to the \onto{isIn} property, \eg \onto{Corridor1}, that we store in a variable named \texttt{robotLoc}.
Then, it shows how to find the OWL classes that classify \texttt{robotLoc} such that they are a leaf in the TBox.
Within Lines~\ref{ln:robotDs}--\ref{ln:robotLoc} the robot location is retrieved by assuming in the ontology that only an \onto{isIn} axiom involving \onto{Robot1} exists.
Lines~\ref{ln:locDescr1}--\ref{ln:locDescr2} initialise a descriptor grounded on \texttt{robotLoc} that maps OWL classes concerning the \onto{Type} expression.
Line~\ref{ln:build} builds the \onto{Type} of \texttt{robotLoc}, \ie it computes a set of descriptors that are grounded in a class representing the robot location (\eg \onto{CORRIDOR}, \onto{LOCATION} and \onto{THING}), and involving \onto{Sub} expressions.
For each class of \texttt{robotLoc}, Line~\ref{ln:ifleaf} checks whether it has only one sub-class, \ie \onto{NOTHING}, which implies that it is a leaf class in the ontology.

The latter example shows that OWLOOP allows to get OWL classes and sub-classes in an OOP fashion, and a similar approach can also be used for the other expressions. 
In particular, at Lines~\ref{ln:build}--\ref{ln:print}, we iterate for each individual to retrieve their class and related sub-classes without using the OWL factory but only relying on the OOP-based internal states of the descriptors.
This is possible thanks to the \texttt{build()} method since it can be used to relate entity sets of different descriptors and, as an example, Lines~\ref{ln:robotLoc}--\ref{ln:locDescr2} show a possible implementation of it.
However, to retrieve \texttt{locIndiv} from \texttt{robot1} directly through building, the descriptor \onto{LinkIndividualDescr} should have been designed to build new individual descriptors concerning the \onto{Type} expression, while it builds descriptors concerning \onto{ObjectLink} and \onto{DataLink} expressions, \ie other \onto{LinkIndividualDescr}.

A simplistic approach could be to always build \emph{full} descriptors for accessing all possible expressions from each \onto{build()} method, but this might impact computation complexity since OWLOOP would synchronise axioms that might be unnecessary. 

\subsection{Compound Descriptor Implementation}
\noindent
In order to highlight the modularity of OWLOOP, Listing~\ref{lst:descrImpl} shows the implementation of a \onto{ClassDescriptor} that inherits the functionalities listed in Section~\ref{sec:soft_func} for two types of expressions, \ie \onto{Instance} and \onto{Sub}.
Indeed, for all the Descriptors based on Table~\ref{tab:mapping}, the same pattern can be used to implement OOP classes encoding OWL knowledge, and that are related among each other based on the ontology, \eg it is possible to \emph{get} or \emph{set} the individuals or the sub-classes of the ground class in an OOP-like manner.
In particular, the get or set entities are not encoded in immutable classes as in the case of the factory design pattern but in OWLOOP descriptors, which can be related to other descriptors through the building methods.

In addition, a developer can exploit OOP-based \emph{polymorphism} to implement a hierarchy of OWLOOP descriptors, which encode string-based identifiers of OWL symbols and procedures to retrieve implicit knowledge based on OWL reasoning in a modular way.
Hence, experts on OWL formalism can build an abstract layer concerned with OWL-related aspects by the means of descriptors, which can be provided to experts in software architecture who aim to use an ontology within an OOP-like paradigm.

For example, Listing~\ref{lst:buildMethod} could be implemented as a method of the descriptor in Listing~\ref{lst:descrImpl}.
In this way, experts in software architectures may rely on an OOP-like method to retrieve the robot position and, at the same time, OWL-formalism experts may exploit a modular interface to the ontology for representing knowledge about the robot position.

\begin{lstfloat}[p]\footnotesize\begin{lstlisting}
public class CDescr extends        ClassGround
                      implements   ClassExpression.Sub<CDescr>,
                                   ClassExpression.Instance<IDescr> {|\Suppressnumber\\[-.6em]|
|\Reactivatenumber|
    private Classes subclasses = new Classes();
    private Individuals individuals = new Individuals(); |\Suppressnumber\\[-.6em]|
|\Reactivatenumber| 
    public CDescr(OWLClass instance, OWLReferences onto) {
        super(instance, onto);    // A constructor based on ClassGround.
    }
    @Override    // Override from ClassGround.
    public List<MappingIntent> readAxioms() {    // Write in the ontology.
        List<MappingIntent> r = Sub.super.readAxioms();
        r.addAll(Instance.super.readAxioms());
        return r;
    }
    @Override    // Override from ClassGround.
    public List<MappingIntent> writeAxioms() {    // Read from the ontology.
        List<MappingIntent> r = Sub.super.writeAxioms();
        r.addAll(Instance.super.writeAxioms());
        return r;
    }|\Suppressnumber\\[-.6em]|
|\Reactivatenumber|
    @Override    // Override from the Sub expression.
    public Classes getsubclasses() {
        return subclasses;    // A factory-based getter of sub-classes names.
    }
    @Override    // Override from Sub, and used in the Sub.build() method.
    public CDescr getsubclassDescriptor(OWLClass cls, OWLReferences onto){
        // Implements an OOP-like getter of sub-classe descriptors.
        return new CDescr(cls, onto);    
    }|\Suppressnumber\\[-.6em]|
|\Reactivatenumber|
    @Override    // Override from the Instance expression.
    public Individuals getIndividuals() {
        return individuals;    // A factory-based getter of instance names.
    }
    @Override    // Override from Instance, and used in Instance.build().
    public IDescr getIndividualDescriptor(OWLNamedIndividual instance, 
                                              OWLReferences onto) {
        // Implements an OOP-like getter of instance descriptors.    
        return new IDescr(instance, onto);    
    }
}
\end{lstlisting}
\caption{%
    The implementation of \texttt{CDescr}: a \texttt{ClassDescriptor} that covers \texttt{Sub} and \texttt{Instance} expressions.
    \texttt{CDescr} is designed to build other instances of \texttt{CDescr} for each OWL class related to the \texttt{Sub} expression, and a \texttt{IDescr} for each OWL individuals related to the \texttt{Instance} expression.
}
\label{lst:descrImpl}
\end{lstfloat}    

\section{Impact}
\label{sec:impact}
\noindent
OWLOOP enables the implementation of modular OOP hierarchies of descriptors that represent specific knowledge in the ontology.
Descriptors can encode both string-based identifiers of OWL symbols, \ie immutable OOP objects, and procedures to exploit implicit knowledge based on polymorphism and OWL reasoning.
Since descriptors can build other descriptors such that they are related to each other based on the ontology, it is possible to exploit OWL-based knowledge through OOP objects that are mutable. 
Moreover, the process of adapting a system when its ontology changes is simplified with OWLOOP because all OWL-related aspects are encapsulated in the descriptors rather than being defined by software components.
Hence, OWLOOP allows to decouple the OWL-based representation of knowledge and its usage within a software system.

OWLOOP is based on aMOR, which is a library providing utilities to use OWL-API.
Since aMOR does not hide the access to OWL-API, developers can keep using any procedure based on OWL-API.
However, other OWL-based APIs (\eg Jena) enable working with different ontology formats, which are currently not compatible with OWLOOP.
Nevertheless, there is no reason for which our OWL to OOP mapping cannot be implemented with other APIs that encompass the OWL formalism.
In addition, other available tools can be used to design an ontology that can be later interfaced with the components of software architectures using OWLOOP descriptors.
Indeed, within the OWL domain, there are no particular features that OWLOOP tackles but other tools do not.
Nonetheless, in the domain of software design, development and maintenance, OWLOOP has features that are not directly provided by other OWL-related APIs, especially when reasoning is concerned.
In particular, OWLOOP descriptors can inherit and extend procedures to evaluate axioms in an ontology by hard-coding only a few string-based identifiers of OWL symbols.
Also, through the building methods, descriptors can access a piece of knowledge without explicitly using any farther axioms. 

Ontologies have been used in a large range of applications, \eg concerning semantic web~\cite{berners2002new}, data access~\cite{xiao2018ontology}, information retrieval~\cite{fernandez2011semantically}, medical knowledge representation~\cite{fung2019knowledge}, internet of things~\cite{gubbi2013internet}, robotics~\cite{olszewska2017ontology}, and ambient assisted living~\cite{rafferty2017activity}. 
Usually, OWL ontologies are deployed in software architectures by assuming that they only contain knowledge that does not change over time.
Other systems exploit ontology structures where individuals are dynamic, while classes and properties are static.
However, some applications also require changing OWL classes over time, \eg to learn scene classification from images~\cite{oldSit}.
In that work, which has been deployed in a software architecture to address human supervision~\cite{sit}, and to classify scenes in a long-term interaction~\cite{memorySit}, we observed that OWLOOP allows to develop a more modular architecture than OWL-API since it can decouple the procedures related to the knowledge representation and its deployment.
For complex systems involving a robot, OWLOOP has been integrated\footnote{A scenes learner implemented with OWLOOP and injected in the aRMOR service can be found at \url{https://github.com/TheEngineRoom-UniGe/ARMOR_OWLOOP}.} in the aRMOR service~\cite{armor_ws}, which allows to use ontologies in the Robot Operating System (ROS).
Furthermore, by investigating the problem of human activity recognition~\cite{s2018Arianna}, we observed that OWLOOP is efficient for system prototyping and evaluation of activity models designed by domain experts.

\section{Conclusions and Future Work}
\label{sec:concl}
\noindent
OWLOOP API performs a passive OWL to OOP mapping that allows to use knowledge inferred from OWL reasoners within the OOP paradigm.
We presented OWLOOP descriptors, which are modular Java classes  interfacing a hierarchy of OOP objects with the knowledge structured in memory as an ontology.
Descriptors enable designing general-purpose interfaces between ontologies and components of software architectures such that drawbacks in the computation are avoided.
In addition, they encapsulate boilerplate code to simplify the development and maintenance of an architecture.
OWLOOP is compatible with other tools to design ontologies, and it is suitable for integrating dynamic ontologies within complex software architectures.

The current version of OWLOOP concerns only the expressions shown in Table~\ref{tab:mapping}, which do not encompass the whole OWL axioms.
For instance, it cannot represent classes through structured disjunctions and conjunctions of class restrictions as defined by OWL.
Instead, it considers all the restrictions to be in conjunction without a specific order. 
This is not a limitation for all the applications where the definition of OWL classes do not change dynamically, because the ontology can be created with other tools and be interfaced with OWLOOP afterwords.
Although there are no reasons for which our OWL to OOP mapping cannot consider all OWL axioms, we present this preliminary implementation to propose our approach. 
Nevertheless, in the future, we want to modularly extend OWLOOP to support all OWL axioms by designing new types of OWLOOP entities and related descriptors.

\section*{Declaration of Competing Interest} 
\noindent
The authors declare that they have no known competing financial interests or personal relationships that could have appeared to influence the work reported in this paper.

\end{document}